\documentclass[letterpaper]{article}
\usepackage{aaai20}
\usepackage{times}
\usepackage{helvet}
\usepackage{courier}
\usepackage[hyphens]{url} 
\usepackage{graphicx,subcaption}
\usepackage{amssymb} %

\newcommand{\frameworkfull}{POpularity-Reinforced Learning for inspired Headline Generation}
\newcommand{\framework}{PORL-HG}

\begin{document}
\title{Attractive or Faithful? Popularity-Reinforced Learning \\for Inspired Headline Generation}
\author{
%Yunzhu Song, Honghan Shuai, Sunglin Yeh, Yilun Wu, Lunwei Ku, Wenchih Peng
Yun-Zhu Song,\textsuperscript{\rm 1}
Hong-Han Shuai,\textsuperscript{\rm 1}
Sung-Lin Yeh,\textsuperscript{\rm 2}
Yi-Lun Wu,\textsuperscript{\rm 1}
Lun-Wei Ku, \textsuperscript{\rm 3}
Wen-Chih Peng, \textsuperscript{\rm 1}\\
\textsuperscript{\rm 1}National Chiao Tung University, Taiwan
\textsuperscript{\rm 2}National Tsing Hua University, Taiwan\\
\textsuperscript{\rm 3}Academia Sinica, Taiwan\\
\{yunzhusong.eed07g,hhshuai,w86763777.eed08g\}@nctu.edu.tw, \\ \{ff936tw\}@gapp.nthu.edu.tw, lwku@iis.sinica.edu.tw, wcpeng@g2.nctu.edu.tw}
\maketitle
\begin{abstract}
With the rapid proliferation of online media sources and published news, headlines have become increasingly important for attracting readers to news articles, since users may be overwhelmed with the massive information. In this paper, we generate inspired headlines that preserve the nature of news articles and catch the eye of the reader simultaneously. The task of inspired headline generation can be viewed as a specific form of Headline Generation (HG) task, with the emphasis on creating an attractive headline from a given news article. To generate inspired headlines, we propose a novel framework called \frameworkfull\ (\framework). \framework~exploits the extractive-abstractive architecture with 1) Popular Topic Attention (PTA) for guiding the extractor to select the attractive sentence from the article and 2) a popularity predictor for guiding the abstractor to rewrite the attractive sentence. Moreover, since the sentence selection of the extractor is not differentiable, techniques of reinforcement learning (RL) are utilized to bridge the gap with rewards obtained from a popularity score predictor. Through quantitative and qualitative experiments, we show that the proposed \framework~significantly outperforms the state-of-the-art headline generation models in terms of attractiveness evaluated by both human (71.03\%) and the predictor (at least 27.60\%), while the faithfulness of \framework~is also comparable to the state-of-the-art generation model.
\end{abstract}

%\noindent
\section{Introduction}

\textit{``A good basic selling idea, involvement and relevancy, of course, are as important as ever, but in the advertising din of today, unless you make yourself noticed and believed, you ain't got nothing''}
\par\hfill--- Leo Burnett \textit{(1891-1971)} \\
Nowadays, users are overwhelmed with rapidly-increasing number of articles from not only news websites but also social media. Therefore, headlines have become more and more important for attracting readers to articles. Articles with eye-catching headlines often attract more attention and receive more views or shares, which is important to content providers since the number of views can be monetized with AD networks and bring revenue. To improve the view rate, some content farms use clickbait headlines to attract users, e.g., ``15 tweets that sum up married life perfectly. (number 13 is hilarious)''. Nevertheless, the clickbait approaches, though effective at the beginning, make users feel annoyed and eventually reluctant to read anything from these websites. It is important to generate attractive headlines while still being faithful to the content.

Headline generation can be regarded as a branch of the article summarization tasks and categorized into extractive-based methods and abstractive-based methods. Extractive-based methods generate headlines by selecting a sentence from the article \cite{higurashi2018extractive}. In contrast, abstractive methods generate the headline by understanding the article and summarizing the idea in one sentence \cite{takase2016neural,hayashi2018headline}. These two kinds of approaches both generate faithful headlines that help readers understand the content at the first glance. However, most existing headline generation approaches do not take the attractiveness into consideration. Zhang et al. observe that interrogative headlines usually attract more clicks and thus formulate the headline generation task as Question Headline Generation (QHG)~\cite{zhang2018question}. Nevertheless, the QHG approach is limited since 1) not every article is suitable for question headlines (e.g., obituaries) and 2) it looks annoying if every headline is in an interrogative form. To the best of our knowledge, this is the first work using the data-driven approach to generate both faithful and attractive headlines in a general form.

However, inspired headline generation introduces at least three new research challenges. First, there are currently only public datasets for the headline generation, and none of them contains information relating to attractiveness, e.g., views, comments, or shares. Second, even with datasets and the extractive-abstractive architecture, it is still challenging to incorporate the attractiveness and faithfulness of information for the extractive-abstractive architecture since i) attractiveness and faithfulness are evaluated based on the sentences rewritten by the abstractor but the gradient can not propagate through the non-differentiable operations of the extractor, and ii) the dependency between extractor and abstractor may lead to slow convergence, i.e., when the extractor is weak and selects a sentence without any popularity-related words, it is difficult for the abstractor to rewrite it into an attractive one. Third, generating headlines based on meaning faithfulness may sometimes conflict with attractiveness. It is challenging to strike a balance between faithfulness and attractiveness.

To tackle these challenges, we present in this paper a framework called \frameworkfull\ (\framework), to generate attractive headlines while still preserving meaning faithfulness to the articles. Specifically, for the first challenge, we build two datasets, \textit{CNNDM-DH} (CNN/Daily Mail-Document with Headline) and \textit{DM-DHC} (Daily Mail-Document with Headline and Comment), based on the \textit{CNN/Daily Mail} dataset \cite{hermann2015teaching,nallapati2016abstractive}, which originally only contains documents with corresponding human written summaries. We further crawl the headlines and headlines with the number of comments for \textsc{CNNDM-DH} and \textsc{DM-DHC}, respectively.\footnote{Since CTR (click-through-rate) is only accessible for news platform owners, we use the comments as the popularity information to train our model, which has been proved to be highly related to CTR \cite{kuiken2017effective}. The details of the datasets are discussed in Section Corpus.} Based on the datasets, we build a state-of-the-art popularity predictor~\cite{lamprinidis2018predicting} to provide the popularity information for unlabeled data. 

Moreover, for the second challenge, we propose a new learning framework that exploits policy-based reinforcement learning to bridge the extractor and abstractor for propagating the attractiveness and faithfulness to the extractor. Moreover, to enhance the ability of the extractor for selecting the sentences containing popular information, we propose Popular Topic Attention (PTA) to incorporate the topic distributions of popular headlines as the auxiliary information. For the third challenge, we design a training pipeline to elegantly strike a balance between attractiveness and faithfulness to avoid generating clickbait-like headlines. Experimental results of the qualitative and quantitative analyses show that \framework ~clearly attracts people's attention and simultaneously preserves meaning faithfulness. The contributions of this paper are summarized as follows.
\begin{itemize}
    \item Instead of the traditional headline generation, we propose the notion of inspired headline generation. To the best of our knowledge, this is the first work utilizing deep learning for generating headlines that are both attractive and faithful in \textbf{general form}. Moreover, the datasets will be released as a public download for future research.
    \item We introduce the \framework, which adopts extractive-abstractive architecture. To incorporate the information of attractiveness and faithfulness for the extractor, we utilize the topic distributions of popular articles as auxiliary information and design a popularity-reinforced learning method with a training pipeline to strike the balance between attractiveness and faithfulness.
    \item The experimental results from both the user study and real datasets manifest that \framework~significantly outperforms the state-of-the-art headline generation models in terms of the attractiveness evaluated by both human (71.03\%) and classifier (27.60\%), while maintaining the relevance compared with the state-of-the-art headline generation models.
\end{itemize}

\section{Related work}
\label{sec:prilim}
\subsection{Headline Generation}
The headline generation task can be seen as a variant of the summarization task with one or two sentences, which is standardized in the DUC-2004 competitions \cite{over2007duc}. Traditional summarization works are often statistical-based and mainly focus on extracting and compressing sentences \cite{knight2000statistics,Cohn2008}. Recently, with the large-scale corpora, many works have exploited neural networks \cite{rush2015neural,filippova2015sentence,cheng2016neural} to summarize articles via data-driven approaches. On the other hand, abstractive method generates summaries or headlines based on document comprehension, which can be considered as a machine translation task. Nallapati et al. propose several novel models to address critical problems, such as modeling keywords and capturing the sentence-to-word hierarchy structure \cite{nallapati2016abstractive}. Since supervised learning often exhibits the ``exposure bias'' problem, reinforcement learning is also used for both of abstractive summarization \cite{paulus2018a,chen2018fast} and extractive summarization \cite{narayan2018ranking}. In addition, Narayan et al. generate extreme news summarization by creating an one-sentence summary answering the question ``What is the article about?''~\cite{NarayanExtreme2018}. However, none of the existing approaches has considered generating attractive headlines with data-driven approaches. Zhang et al. formulate the attractive headline generation task as QHG (Question Headline Generation) \cite{zhang2018question}, according to the observation that interrogative sentences attract more clicks. However, in spite of the effectiveness achieved by a question form headline, it is still not suitable to generate every headline as a question.
% The pointer network \cite{vinyals2015pointer} and coverage mechanism \cite{see2017get} have been recently introduced to generate accurate details and to prevent word repetition.

\subsection{Popularity Prediction}
Beside information faithfulness, the effectiveness of the headlines, i.e. click-through-rate, is also important as mentioned above. Kuiken et al. analyze the relationship between CTR and the textual/stylistic features of millions of headlines, which can provide insights for how to construct attractive headlines \cite{kuiken2017effective}. Meanwhile, the online news on social media has also been analyzed by identifying the salient keyword combinations and recommending them to news editors based on the similarity between popular news headlines and keywords \cite{weng2018recommendation,szymanski2017helping}. To predict the popularity of a headline, Lamprinidis et al. use an RNN with two auxiliary tasks, i.e., POS tagging and section prediction~\cite{lamprinidis2018predicting}. Nevertheless, none of the existing approaches generate headlines in terms of popularity prediction.
%Kourogi et al. first analyze the relationship between the popularity of news articles published on traditional media, and Twitter-focused news headlines posted by the official Twitter account, and then learn to suggest headlines attractive to users by means of hand-crafted features \cite{kourogi2015identifying}.

\begin{table}[!htbp]
    \caption{Dataset information}
    \label{tab:dataset}
    \centering
    \begin{tabular}{llll}
    \hline
             & Train & Val & Test \\
    \hline
         CNNDM-DH  & 281208   & 12727   & 10577 \\
         DM-DHC    & 138787   & 11862   & 10130\\
    \hline       
    \end{tabular}
\end{table}

\section{Corpus}
\label{sec:corpus}
\subsection{Dataset and Headline Performance Analysis}
\label{sec:dataset}
To automatically generate a headline that is not only faithful but also eye-catching, the summarization dataset and the popularity statistical information are needed. Therefore, we build the datasets \textit{CNNDM-DH} (CNN/Daily Mail-Document with Headline) and \textit{DM-DHC} (DailyMail-Document with Headline and Comment) based on \textit{CNN/Daily Mail} dataset \cite{nallapati2016abstractive,hermann2015teaching}, which contains online news articles paired with multi-sentence summaries without headlines. Hence, we access the original online news pages to crawl the headlines and popularity information for both CNN and the Daily Mail, and then remove the damaged data. Detail information is shown in Table~\ref{tab:dataset}.

Since only the comment counts and share counts are available for DM-DHC datasets, we first validate the idea of using them as the popularity scores for training. Following the previous research \cite{kuiken2017effective}, which studies the relationship between ``clickbait features'' and CTR by extracting features from headlines to form 11 null hypotheses whose significance were examined using non-parametric Mann-Whitney $U$ test\footnote{Non-parametric Mann-Whitney $U$ test is widely-used for the significance test of non-normal distributions.}. We use the same null hypotheses to determine whether the significance of the CTR, comment counts, and share counts are similar. The result shows that the significance tests of the CTR and comment counts are almost the same (7/8), while the significance tests of the CTR and share counts exhibit more difference (6/8). Therefore, we use the comment counts as the ground truth of attractiveness for training and testing.\footnote{The details of the analysis and datasets are available at https://github.com/yunzhusong/AAAI20-PORLHG.}

%% section 4
\section{\frameworkfull}
\label{sec:porl-hg}
\begin{figure}[t]
\centering
\includegraphics[width = .45\textwidth]{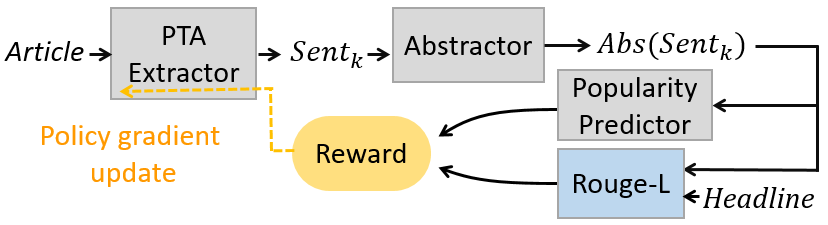}
\caption{The framework of \framework.}
\label{fig:model}
\end{figure}

To strike a balance between faithfulness and attractiveness, we propose a novel framework called \frameworkfull\ (\framework). The framework is shown in Figure \ref{fig:model}. Specifically, \framework~first exploits a hybrid extractive-abstractive architecture to generate headline effectively, i.e., an extractor selects a candidate sentence from the article and an abstractor then rewrites the headline based on the candidate sentence. To provide the faithfulness and attractiveness information for the extractor, inspired by the pointer network~\cite{vinyals2015pointer}, we propose Popular Topic Attention (PTA) by utilizing the topic distribution of the related and popular headlines. Moreover, to train the abstractor for writing eye-catching headlines, we build a headline popularity predictor by using CNN to extract features from headlines and an LSTM to predict the popularity score. The popularity score is integrated into the loss function of the abstractor to encourage the generation of attractive headlines and preserving the faithfulness simultaneously. However, this basic approach suffers from the issue that the extractor is not trained with the popularity score to select attractive sentences. Therefore, the reinforcement learning (RL) techniques are used to train the extractor. In the following, we introduce each module in \framework, and present the training pipeline afterward.

%% Figure 2 for sec 4.4

\subsection{Inspired Extractor}
\label{sec:extractor}
\begin{figure*}[t]
\centering
\includegraphics[width = .95\textwidth]{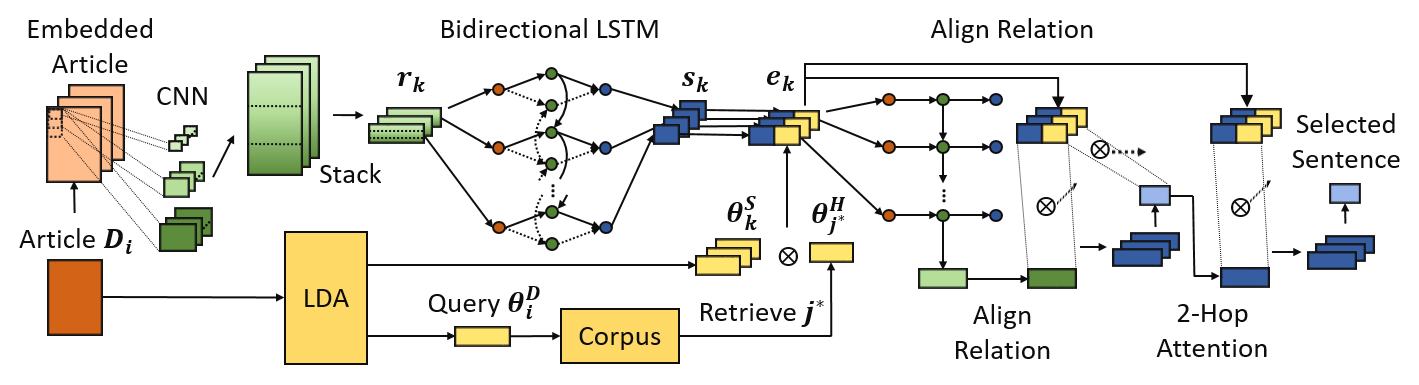}
\caption{Extractor agent with PTA: a CNN is used as the sentence representation encoder, and sentence embedding $r_k$ is further encoded by the bidirectional LSTM to acquire the context-aware sentence representation $s_k$. The popular topic distribution $\theta^H_{j^{\ast}}$ related to the input article is multiplied with the topic distribution of sentences and concatenated with the $s_k$. Then, the last state of $e_k$ is fed to the pointer for guiding the sentence selection.}
\label{fig:extract}
\end{figure*}
Given an article, the goal of the extractor is to choose a salient sentence from the article for the following rewriting task of the abstractor, which requires 1) the sentence representation and 2) the ground truth of the salient sentence for training. For the sentence representation, we first embed each word by word2vec and concatenate the word embeddings for each sentence in the article. Afterward, we exploit a convolutional neural network \cite{kim2014convolutional} with different kernel sizes to capture the complete semantic meaning of each sentence, which is denoted as ${r}_k$ for the $k$-th sentence in the article. However, the long-term relationship between sentences is not captured for generating the sentence embeddings. Therefore, a bidirectional LSTM is then applied to improve the embedding representation ${r}_k$, which is denoted as ${s}_k$. For the ground truth of the salient sentence, we use a proxy label by calculating \textit{ROUGE-L} score of each sentence in the article \cite{chen2018fast,nallapati2017summarunner} and marking the highest one as the proxy training label.

After deriving the sentence embedding and proxy label, one basic approach is to use existing extractive summarization models for selecting a sentence. However, since the proxy label only considers the faithfulness (\textit{ROUGE-L}), the selected sentence may not contain any attractive keywords, which makes the rewriting task of the abstractor difficult. To solve this issue, we propose Popular Topic Attention (PTA) that uses attention with the popularity information as a pointer to select a sentence from the article. Specifically, we first exploit Latent Dirichlet Allocation \cite{blei2003latent} to generate the topic distributions of articles and headlines. Let $\theta_{i}^{D}$ and $\theta_{i}^{H}$ denote the topic distributions of $i$-th article $D_i$ and the corresponding headline $H_i$, respectively. Afterward, for input article $D_i$, we calculate the inner product of $\theta_{i}^{D}$ and $\theta_{j}^{D}$ ($\forall j\neq i$), and retrieve the top-$m$ similar articles. Among the top-$m$ similar articles, the most popular article $D_{j^{\ast}}$ is selected and the corresponding topic distribution of the headline, i.e., $\theta^H_{j^{\ast}}$, is used as a reference. For each sentence $s_k$, we use $\theta^S_k\otimes \theta^H_{j^{\ast}}$ as the popularity information, where $\otimes$ denotes the element-wise multiplication. The operation preserves the topics appeared in both of $\theta^S_k$ and $\theta^H_{j^{\ast}}$, allowing the model to find the topics that are both faithful and attractive.

To provide both faithfulness and attractiveness information for sentence selection, for each sentence $s_k$, PTA constructs $e_k$ by concatenating $s_k$ and $\theta^S_k\otimes \theta^H_{j^{\ast}}$, i.e.,
$$e_k = [s_k;\theta^S_k\otimes \theta^H_{j^{\ast}}],$$
where [$\cdot$;$\cdot$] denotes the concatenation. Then, the sentence information $e_k$ will be fed into the sentence selector to make the selection, which is an LSTM equipped with 2-hop attention. The first attention step is to get the context vector $c$ by applying the glimpse operation to every sentence $e_k$ as follows:
$$    u_k =\nu^T_g tanh(W_{g_1}e_k + W_{g_2}z) $$
$$    \alpha _k= softmax(u_k)$$
$$    c = \sum_{k}\alpha _k W e_k $$ 
where $z$ is the initial state of the LSTM and $W$, $W_{g_1}$, $W_{g_2}$ and $\nu_g$ are all trainable parameters, and $\alpha_k$ is the attention coefficient of $k$-th sentence for deriving $c$. After this, the second attention step is to attend $e_k$ again by the context vector $c$, which results $o_k$, i.e.,
$$    o_k = \nu^T_p tanh(W_{p_1}e_k + W_{p_2}c),$$
where $W_{p_1}$, $W_{p_2}$ and $\nu_p$ are also trainable parameters for deriving the sentence selection probability. Let $o$ denote the output vector composed of $o_k$ for each sentence $k$. Finally, the extraction probability $P(k)$ of extracting the $k$-th sentence can be obtained from $o$ by using the softmax function as
\begin{equation}
    P(k) = softmax(o).
    \label{eqn: P(j)}
\end{equation}
Figure \ref{fig:extract} shows the network architecture of the inspired extractor with the PTA mechanism. Specifically, this approach differs from the previous bidirectional LSTM that generates sentence representation, we add a new mechanism to encourage the model to select a sentence that is not only informative but also eye-catching.

\subsection{Inspired Abstractor}
%\begin{figure}[t]
%\centering
%\includegraphics[width = .45\textwidth]{Model_abstractor.png}
%\caption{The abstractor architecture.}
%\label{fig:abs}
%\end{figure}

The abstractor compresses and paraphrases an extracted article sentence to a headline sentence, for which, we use the standard encoder-aligner-decoder with attention mechanism. To deal with the out-of-vocabulary (OOV) words, we apply a copy mechanism \cite{see2017get}, which can directly copy words from the article. Additionally, to make the headline more eye-catching, the rewriting ability is also important. One basic approach of training the abstractor for generating attractive headlines is to construct a corpus containing multiple headlines with the corresponding popularity scores for the same article. However, deriving such datasets costs highly. In addition, the distribution of popularity scores (comment counts) follows a long-tailed distribution, which makes the prediction biased. Therefore, we transform the popularity score into a binary label, where $0$ represents the popularity score which is smaller than the median and $1$ otherwise. Next, we pre-train the binary classifier by using headlines, which are encoded with the same CNN used in the extractor. Then, we classify the results through an LSTM. The classification score will be returned to the abstractor as an auxiliary loss or reward. 

% More details of the pretraining are deliberated in Sec. \ref{sec:cls}.
\iffalse
\subsection{Popularity Predictor}
\begin{figure}[t]
\centering
\includegraphics[width = .45\textwidth]{Model_classifier.png}
\caption{The classifier architecture.}
\label{fig:cls}
\end{figure}
\fi
\subsection{Training Pipeline}
When training the \framework, the gradient derived in the abstractor cannot be propagated back to the extractor. In order to perform an end-to-end training and to combine the popularity information, we apply the reinforcement learning with the standard policy gradient to connect the extractor, abstractor and auxiliary classifier. It is worth noting that training from a random initialization is difficult due to the dependent interplay between the extractor and abstractor, e.g., the abstractor cannot learn rewriting an attractive sentence when the extractor is not well-trained and thus selects meaningless sentences. Moreover, an abstractor without a good rewriting ability leads to a noisy estimation of the standard policy gradient, which deteriorates the training of the extractor. Hence, pre-training the abstractor, extractor and classifier before starting the reinforcement learning is necessary.

\subsubsection{Extractor Training}
The task for the inspired extractor is to select an essential and eye-catching sentence from the article. Since most of the headline generation dataset does not include the extracted headline labels, we offer the proxy label similar to \cite{nallapati2017summarunner} for the extractive summarization task. The label is acquired by calculating \textit{ROUGE-L} score for every sentence with respect to the ground-truth headline $H_i$. That is, the proxy target label $y_i$ for the $i$-th article $D_i$ is obtained by $y_i = argmax \left( ROUGE\textit{-}L_{recall}(D_i, H_i) \right)$, and the loss function of the extractor is:
\begin{equation}
     L_{ext} = -\frac{1}{N}\sum_{i=1}^{N} y_i log(P(y_i)).
     \label{Eq:proxylabel}
\end{equation}
In addition to the traditional classification label for faithfulness, we propose a new pre-trained proxy label $y'_i$ taking the popularity information into consideration. Specifically, since the topic distribution set $\textbf{e}$ represents the similarity between each sentence and the retrieved popular headline, the summation of topic values can be viewed as a popularity score $\sum{e_j}\in \mathbb{R}^{1}$. We normalize the summation by subtracting the mean and dividing
the variance, then choosing the sentence with maximum value to be one of the extraction label $y'_i$, i.e., $y'_i=argmax(normalize(\sum{e_1},...,\sum{e_N}))$. The final loss of the extractor, denoted as $L_{ext}^{\prime}$, is derived as follows:
\begin{equation}
  L_{ext}^{\prime} = L_{ext}-\frac{1}{N}\sum_{i=1}^{N} y'_i log(P(y'_i)). 
  \label{Eq:newproxylabel}
\end{equation}

\subsubsection{Abstractor Training}
The training data for the abstractor are pairs of extracted proxy headline $h_{gen}$ (obtained from Eq.\ref{Eq:proxylabel}) and the ground-truth headline. Specifically, the objective function $L_{abs}$ has two main purposes: 1) to minimize the cross-entropy loss between the extracted proxy headline $h_{gen}$ and ground-truth headline and 2) to increase the popularity score of $h_{gen}$. The objective function is derived as follows: 
$$L_f(\theta_{abs}) = -\frac{1}{M}\sum_{m=1}^{M}{logP_{\theta_{abs}(w_m|w_{1:m-1})}},$$
$$L_a(h_{gen}) = - pop(h_{gen}),$$
$$ L_{abs} = L_f + L_a,$$
where $w_m$ is the $m$-th token in the ground-truth headline and $M$ is the headline length.

\subsubsection{Popularity Predictor Training}
\label{sec:cls}
We train a binary classifier as an auxiliary model. There are two reasons for training the popularity predictor instead of the regression model. The first reason is that predicting the exact comment counts may result in overfitting to the outlier data, and the second reason is that, there is no information about other factors that affect the comment counts for a precise prediction, e.g., events. Therefore, we transform the popularity score into a binary label, where $0$ represents the popularity score as being smaller than the median, otherwise the score is $1$. A state-of-the-art popularity predictor~\cite{lamprinidis2018predicting} is trained to minimize the cross entropy. The final accuracy of popularity predictor is 65.46\% on our test data.

\subsubsection{Reinforcement Learning}
For the purpose of bridging the back propagation and introducing the classifier reward, we perform the RL training to optimize the whole model. We make the sentence extractor into an RL agent. For every extraction step, the agent observes the current state $s=(D, \theta^D,$\boldmath${\theta}^S$\unboldmath $)$, where $D$ is the article, $\theta^D$ is the topic distribution of article and \boldmath $\theta^S$ \unboldmath is the set of topic distributions for each sentence. After that, if the agent takes the action $j$, i.e.,
$$j \sim \pi (s) = P(j),$$
where $P(j)$ is from Eq.\ref{eqn: P(j)}, it means that the agent selects $j$-th sentence from the article under the current policy $\pi$. The abstractor then rewrites the selected sentence and send it to the popularity predictor. Finally, the agent receives the reward $r$ by adding (1) the ROUGE-L score between the target sentence and the rewritten sentence and (2) the score of the rewritten sentence from popularity predictor, i.e.,
$$r = ROUGE\textit{-}L_{F_1}(abs(s_j), H) + pop(abs(s_j)).$$ 
Moreover, due to the high variance of the vanilla policy gradient~\cite{williams1992simple}, we add another mechanism, the Advantage Actor-Critic (A2C)~\cite{mnih2016asynchronous} to stabilize the training process.

It is worth noting that the role of the PTA is to guide the optimization of RL. Specifically, RL randomly selects sentences to explore the action space at the early training stage, which makes the training difficult. Without a good abstractor, RL can only receive little reward from the popularity predictor, which also makes the training of the extractor difficult. With the help of the PTA, the extractor can select a better sentence at the early training stage.

\section{Experimental Results}
\label{sec:exp}
We conduct the qualitative and quantitative experiments with two real datasets
%\footnote{The two real datasets is released.} 
to evaluate \framework. For the qualitative evaluation, we provide the case study of generated headlines, and conduct a user study via asking users to evaluate the attractiveness, relevance, and grammaticality of the headlines generated by different approaches. For the quantitative evaluation, we compare different approaches in terms of attractiveness and faithfulness. To evaluate the attractiveness of generated headlines, we show the average score derived from the state-of-the-art popularity predictor~\cite{lamprinidis2018predicting}. Moreover, we analyze the features of the popularity hypotheses mentioned in Sec. Corpus, provided along with the source code. To evaluate the faithfulness, we report the ROUGE scores for different approaches. In addition, we show the training reward curve of~\framework~with and without the popularity information, and analyze the attention of the CNN features.
\begin{table*}[h]
\small
    \caption{Examples from the testing data showing the ground-truth headline and two generated headlines.}
    \label{example}
    \begin{tabular}{p{5cm}p{5.5cm}p{6cm}}
        \hline
        Ground-truth Headline & Chen et al. & \framework \\
        \hline
        \hline
        %Feral cat pictured killing and eating a wallaby 
        %& Is this the feral cat killing and eating a four-kilogram marsupial? 
        %& Is this the most threats of the country? Feral cat killing and eating a four-kilogram marsupial \\
        Come rain or hail! Surfers hit the southern California coast as freezing showers turn the beach white
        & Surf city's surf city transformed into a white canvas ... but didn't stop the surfers from hitting the shore 
        & Beach! California beach transformed into white canvas after dumping of hail - but didn't stop surfers from hitting the shore \\
        \hline
        From Rihanna to Madonna, new trend features designs of weapons, drugs and body parts 
        & Madonna's new black crocodile handbag causes stir for drug-related slogan 
        & The worst offenders in the luggage? Designers create naked women and abandoned babies into their ranges \\
        \hline
         Netflix announce planet earth sequel our planet for 2019 
         &  Now, the sweeping documentary series `planet earth' is getting a sequel  
         & `Planet earth' is getting a sequel, says sweeping documentary \\
        \hline
    \end{tabular}
\end{table*}
\subsection{Baselines}
We implement the following baseline models and conduct an ablation task. Following the setting of \cite{chen2018fast}, only an upper bound of the headline length is set (30 tokens) for the learning-based models.

\begin{itemize}
  \item \textbf{IR\textsubscript{BM25}} is a bag-of-words retrieval function. It indexes every headline in the training corpus, and feeds documents to search for the best matching headline as output according to the BM25 relevant score function.
  \item \textbf{Random} selects a sentence randomly from the article.
  \item \textbf{PREFIX} takes the first sentence as the headline.
  %\item \textbf{proxy-label} is the target for pretraining extractor, which is determined by maximizing the ROUGE score. Therefore, it is the upper bound for the extractive task.
  \item \textbf{Seq2Seq} employs a bidirectional LSTM as the sentence encoder, and use another bidirectional LSTM to obtain article level representation. A two-layer LSTM is then applied to decode the article representation. All models are equipped with the attention mechanism.
  \item \textbf{Chen et al.~\cite{chen2018fast}} pre-train the extractor to minimize the cross-entropy loss, while the target is the proxy label. Then, they apply RL to train the extractor and use an abstractor to rewrite the sentence. Accordingly, the training target is the ground-truth headline.
   \item \textbf{See et al.~\cite{see2017get}} uses the pointer network and coverage mechanism to generate headlines.
\end{itemize}

\subsection{Qualitative Results}
%% Human evaluation result
%% Example
To better understand what can be learned by our model, Table~\ref{example} shows the ground-truth and generated headlines from test data as a case study. \framework~can generate the headline with different forms, including interrogative sentence, exclamatory sentence or quoting the emphasis statement, depending on the suitability. Moreover, the headlines generated by~\framework~sometimes express stronger sentiments (the first and second examples), which may make users feel stronger emotions for the headlines and lead to click. In contrast, without the information of attractiveness, \cite{chen2018fast} only headlines that summarize the articles are generated. Besides, \framework~sometimes uses the eye-catching words at the beginning of the headlines to draw attentions as shown in the first example.

\begin{table}[t]
    \caption{Human evaluation results.}
    \label{attractive human}
    \begin{tabular}{lll}
    \hline                 &  \framework & Chen et al.\\
    \hline Attractiveness  & 236 (63.10\%)    & 138 (36.90\%) \\
    \hline Relevance       & 77 (35.65\%)    & 74 (34.26\%)  \\
    \hline Grammaticality  & 3.95            & 3.64  \\
    \hline
    \end{tabular}
    \label{human task}
\end{table}

To evaluate the performance of inspired headline generation, we conduct a user survey with 73 users, where 32 participated users have research experience in NLP/deep learning and the rest of the users are not familiar with NLP/deep learning. For the human evaluation, we consider the following three modalities. 1) Attractiveness: given two headlines generated by PORL-HG and \cite{chen2018fast}, we ask users to choose the one that he/she would click; 2) Relevance: given the human written summary that provided in the CNN/Daily Mail Dataset and the headlines generated by different approaches, users are asked to answer whether the headlines are related to the given summary and can select \textit{more than one headline} as relevance; and 3) Grammaticality: given generated headlines, people are asked to rate the generated headlines from 1 to 5 (the higher score indicates a better result). 
%The experiment details of user survey is in the supplementary.

Table~\ref{human task} shows the attractiveness, relevance and grammaticality of headlines generated from the state-of-the-art method and \framework. The results manifest that 63.1\% users think that the headlines generated by \framework~ are more attractive, while only 36.9\% users think that the headlines generated by~\cite{chen2018fast} are more attractive. For the relevance, the scores of \framework~and~\cite{chen2018fast} are close, indicating that \framework~generates headlines with higher attractiveness without affecting the faithfulness compared with the state-of-the-art method. Furthermore, the user survey shows that the grammatical quality of \framework\ is slightly better. By our observation, the reason might be the readability of shorter words. The average token length of \framework\ and \cite{chen2018fast} are 5.15 and 5.47 respectively.
%provides different forms with eye-catching words, e.g., \textit{"Beach!"} in the first example of Table \ref{example} with an exclamatory form, and 

\subsection{Quantitative Results}
We statistically analyze the attractiveness and faithfulness of the headlines. First, the attractiveness is evaluated by the state-of-the-art popularity predictor~\cite{lamprinidis2018predicting} and reports the percentage of headlines classified as attractive. Table \ref{tab:attractiveness} shows that \framework~ achieves the best attractiveness.
Second, we follow previous works \cite{nallapati2017summarunner,see2017get,zhang2018question}, and use ROUGE \cite{lin2004rouge} as a metric to evaluate the faithfulness. Table \ref{auto metrics} is split into extraction and abstraction results for clear comparisons. For the extraction, the performances of IR$_{BM25}$ and random models are poor, which suggests that memorizing the whole training corpus or randomly picking does not work. In contrast, the \textit{PREFIX} model performs quite well, which is expected given that most news articles state the key points in the first sentence. We implement an extractor (denoted as \textit{ext}) with a simple pointer \cite{vinyals2015pointer}, which performs slightly worse than \textit{PREFIX}. The extraction result of \framework\ shows that the way we incorporate the popularity information does not affect the ROUGE score. It is worth noting that the abstraction result of \framework\ is slightly smaller than \cite{chen2018fast} in terms of ROUGE scores because the proposed PORL-HG adopts few more different words to make headlines attractive, which can be observed from the copy rates, i.e., 95.30\% and 96.97\% for \framework\ and \cite{chen2018fast}, respectively. Third, Table~\ref{auto metrics attract} shows the ablation task of our model evaluated by Meteor and Attractiveness. The Meteor metric is used for faithfulness comparison, which is calculated by the recall and the precision, and takes synonyms into consideration. The \framework\ significantly outperforms the baseline by 27.60\% for attractiveness while slightly improves the baseline in terms of Meteor. Moreover, the PTA contributes the most for attractiveness (from 40.76 to 44.06), suggesting that simultaneously considering topic distributions of sentences and popular headlines is effective.

\begin{table}[t]
\begin{center}
\caption{Attractiveness evaluation results.}
\label{tab:attractiveness}
    \begin{tabular}[t]{|l|c|}
    \hline
    Models & Attractiveness\\
    \hline
    Ground truth & 24.53 \\
    Seq2Seq & 33.29 \\
    Chen et al. & 34.53  \\
    See et al. & 37.63 \\
    \framework & \textbf{44.06} \\
    \hline
    \end{tabular}
\end{center}
\end{table}

\begin{table}[t]
\caption{Faithfulness evaluation results, where CP denotes the copy rate. Note that the goal of \framework~is to ``maintain'' the faithfulness and ``improve'' the attractiveness instead of improving the faithfulness and attractiveness simultaneously.}
\label{auto metrics}
    %\begin{tabular}[t]{|l|llll|}
    \begin{tabular}{|p{0.35\columnwidth}|p{0.1\columnwidth}p{0.1\columnwidth}p{0.1\columnwidth}p{0.1\columnwidth}|}
    \hline
    Models & R-1 & R-2 & R-L & CP\\
    \hline
    \multicolumn{5}{|c|}{Extraction results}\\
    \hline
    IR\textsubscript{BM25} & 10.29 & 1.80  &  8.11 & -\\
    Random & 13.74  & 2.92 & 11.04  & - \\
    PREFIX & 32.36  & 13.66  & 26.67 & - \\
    ext     & 32.29 & 13.65  & 26.48  & - \\
    PORL-HG w/o abs  & \textbf{32.51} & \textbf{13.79} & \textbf{26.82} & - \\
    \hline
    
    \hline
    \multicolumn{5}{|c|}{Abstraction results}  \\
    \hline
    Seq2Seq & 13.57  & 3.07 & 11.82  & 66.59\\
    See et al. & 27.80  & 12.06 & 23.30 & 97.00\\
    %ext+abs+RL$^\star$ & 34.84  & 15.91  & 30.26 \\
    Chen et al. & \textbf{34.84}  & \textbf{15.91}  & \textbf{30.26} & 96.97\\
    \framework & 34.23 & 15.35 & 29.48  & 95.30\\
    \hline
    \end{tabular}
\end{table}

\begin{table}[t]
\caption{Ablation study evaluation results.}
\label{auto metrics attract}
    \begin{tabular}[t]{|l|ll|}
    \hline
    Models & Attractiveness & Meteor\\
    \hline
    Chen et al. & 34.53  & 17.28   \\
    w/o pop, topic loss & 40.76  & 17.25  \\
    w/o pop & 40.96  & 17.53  \\
    \framework & \textbf{44.06} & \textbf{18.07}  \\
    \hline
    \end{tabular}
\end{table}

\begin{figure}[t]
\centering
  \begin{subfigure}[t]{0.49\linewidth}
    \centering\includegraphics[width=\linewidth]{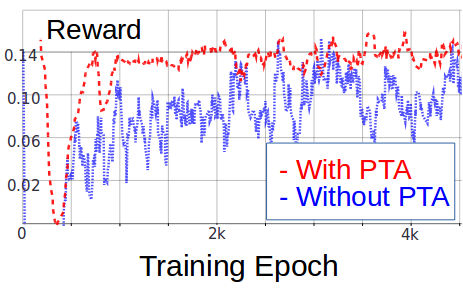}
    \caption{The reward curve during training of \framework.}
    \label{fig:training curve}
  \end{subfigure}
  \begin{subfigure}[t]{0.49\linewidth}
    \centering\includegraphics[width=\linewidth]{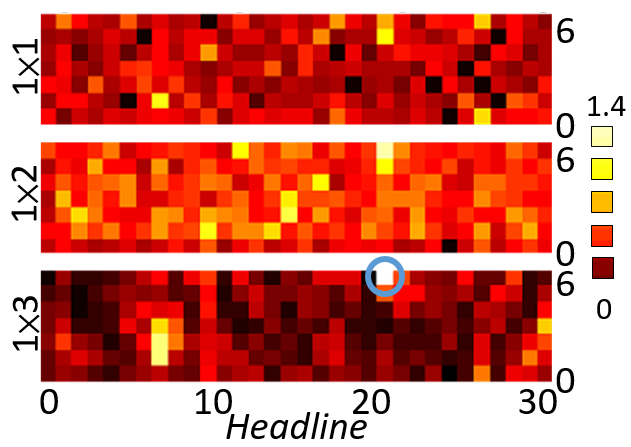}
    \caption{The attention to the feature maps of the CNN.}
    \label{fig:heatmap}
  \end{subfigure}
  \caption{Analysis of the proposed PTA.}
    \label{fig:study}
\end{figure}

To further validate the effect of the proposed PTA, Figure~\ref{fig:training curve} illustrates the training reward curve of \framework\ with and without PTA (red and blue curves, respectively). \framework\ with PTA is more stable and achieves saturation more quickly. This is because RL without PTA randomly selects sentences to explore the action space at the early training stage, which makes the training of the attractive abstractor difficult. Without a good abstractor, RL can only receive a little reward from the popularity predictor, which also makes the training of the extractor difficult. With the help of the PTA, the extractor can select a better sentence to gain the higher reward at the early training stage.

%% Figure

To analyze how the popularity predictor works, we visualize the convolution results for the generated headlines. Figure~\ref{fig:heatmap} shows the intensity of convolution feature map for different kernel sizes. Each column represents the feature map of a headline. From the heat map, we first observe that the features derived by $conv 1 \times 2$ tend to be more attended, meaning that the popularity predictor pays more attention on 2-word level. Moreover, since the headline 21 contains high attention values for $conv 1\times3$ (highlighted by the blue circle), we further investigate the headline, which is ``\textit{Glow-in-the-dark tampons shed light on water pollution}''. The popularity predictor attends on "Glow-in-the-dark tampons shed", which shows the popularity predictor actually focuses on the important term (e.g., proper nouns). Moreover, we can also observe the spotting behavior of attention models for classifying the headline popularity, since the large values for ${conv 1\times1}$ and $conv 1\times2$ are only in one or two positions for a headline.
%``\textit{Splendid kitchen manchester chef creates 2,000-calorie burger}'' and 
\section{Conclusion and Future Work}
\label{sec:conclu}
In this paper, we tackle the challenging task of generating an eye-catching headline with general form. To strike a balance between faithfulness and attractiveness, we propose the \frameworkfull\ incorporating the topic distributions of sentences and popular headlines. We also verify the effectiveness of each module in \framework\ carefully by the quantitative and qualitative experiments, which demonstrate that our model outperforms the state-of-the-art baselines on the attractiveness while simultaneously maintaining the faithfulness. In the future, we plan to explore the possibility of designing a better RL algorithm purely on the abstractive system. 

\section{Acknowledgement}
\label{sec:ack} 
This work was supported in part by the Ministry of Science and Technology of Taiwan under Grants MOST-108-2221-E-009-088, MOST-108-2622-E-009-026-CC2, MOST-108-2634-F-009-006, MOST-108-2221-E-001-012-MY3, MOST-108-2218-E-009-050, and by the Higher Education Sprout Project by the Ministry of Education (MOE) in Taiwan through grant 108W267.

\bibliographystyle{aaai}
\bibliography{900-reference}

\end{document}

\section{Supplementary Material}

\subsection{Corpus}
The \textit{CNN/Daily Mail} dataset contains online news articles (781 tokens on average) paired with multi-sentence summaries. Following the scripts supplied by Nallapati et al. \cite{nallapati2016abstractive}, we obtain the same version of the data; however, headlines are not included. Hence, we use the URL link to access the original online news page and crawl the headlines (22 tokens on average) for both CNN and the Daily Mail, and then remove the data without headlines due to damaged or missing website pages. The dataset information is listed in Table~\ref{tab:my_label}, and can be download in \cite{PORLonline}.

%% Table of dataset for sec 3.1
\begin{table}[!htbp]
    \caption{Dataset information}
    \centering
    \begin{tabular}{llll}
    \toprule
             & Train & Val & Test \\
    \midrule
         CNNDM-DH  & 281208   & 12727   & 10577 \\
         DM-DHC    & 138787   & 11862   & 10130\\
    \bottomrule       
    \end{tabular}
    \label{tab:my_label}
\end{table}

\subsection{Correlation Analysis of CTR, comments and shares}
\label{sec:hypothesis}
%% Table of hypothesis for sec 3.2
\begin{table*}[t]
\caption{List of hypotheses and the corresponding p-value of the significance test, where $\ast$ indicates significant hypothesis (p-value $\leq 0.05$).$^\dag$ Note the p-value of CTR is referenced from Kuiken et al. \protect\cite{kuiken2017effective}.}
\begin{tabular}{p{0.6cm}p{12cm}p{1cm}p{1cm}p{1cm}}
\toprule
\multicolumn{2}{l}{Hypothesis} & CTR$^\dagger$ & Comment & Share\\
\toprule
H1  & Longer headline($>$50 characters) are preferred over shorter headlines  & 0.297 & ${0}^\ast$ & ${0}^\ast$ \\
H2  & Headlines with short words ($\leq$ 7characters per word) are preferred   &${0.024}^\ast$ & ${0}^\ast$ &${0}^\ast$\\
H3  & Headlines containing a question are preferred  & ${0.019}^\ast$ & ${0}^\ast$ & $ {0}^\ast$\\
H4  & Headlines containing a partial quote are preferred over not containing any quote & 0.239 & 0.996 & 0.971\\
H5  & Headlines not containing any quote are preferred over containing full quote  & ${0.03}^\ast$ & 0.848 & 0.111\\
H6  & Headlines that contain one or more signal words are preferred & ${0.002}^\ast$ & ${0}^\ast$ & ${0.001}^\ast$\\
H7  & Headlines that contain one or more personal or possessive pronouns are preferred  & ${0}^\ast$ & ${0}^\ast$ & ${0}^\ast$\\
H8  & Headlines that contain one or more sentimental words are preferred & ${0.018}^\ast$ & ${0}^\ast$ & ${0}^\ast$\\
H9  & Headlines that contain one or more negative sentimental word are preferred  & ${0.001}^\ast$ & ${0.001}^\ast$ & ${0.015}^\ast$\\
H10  & Headlines that contain a number are preferred over headlines that do not & 0.202 & ${0}^\ast$ & ${0.060}^\ast$\\
H11  & Headlines that start with a personal or possessive pronoun are preferred & ${0.002}^\ast$ & ${0}^\ast$ & 0.429\\
\hline
\end{tabular}
\label{table:hypothesis}
\end{table*}
Following the previous research \cite{kuiken2017effective}, we study the relationship between ``clickbait features'' and CTR by extracting features from headlines to form 11 null hypotheses whose significance were examined using non-parametric Mann-Whitney $U$ test. Here, the same null hypotheses are used to determine whether the significance of the CTR, comment counts, and share counts are similar. Table~\ref{table:hypothesis} shows the testing results, suggesting that the significance tests of the CTR and comment counts are almost the same (7/8), while the significance tests of the CTR and share counts exhibit more difference (6/8).

\section{Supplementary Experimental Results}
\subsection{Attractiveness Analysis}
To gain more insights, we analyze how different methods perform regrading to the popularity hypotheses mentioned in Sec. \ref{sec:hypothesis}. Specifically, we transform the hypotheses into features. For example, the hypothesis 1 ``\textit{Longer headline($>$50 characters) are preferred over shorter headlines}'' is converted into \textit{the average character length of a headline}. The transformed features are listed in Table \ref{table:hypo anal}. From our dataset, DHC, there are 8 significant hypotheses and 3 non-significant hypotheses. Interestingly, our headline outperforms GT and the baselines for 7 out of 8 significant features. Although PORL-HG is data-driven and does not add the hypotheses into training, e.g., adding regularization terms, the headlines generated by PORL-HG fit the features of the hypotheses, which means that the potential popularity information can be integrated into the model by the attractive classification rewards. It is worth noting that PORL-HG does not outperform others in terms of the percentage of headlines containing a question mark (H3), which indicates that PORL-HG generates inspired headlines in different forms without fully relying on the interrogative forms.

%% Attractiveness evaluation
\begin{table*}[t]
    \caption{The popularity features. The following 11 features are transformed from the hypotheses stated in Table \ref{table:hypothesis}. GT indicates the abbreviation of ground-truth headlines, and Sig$\star$ is short for significant.}
    \label{table:hypo anal}
    %\begin{tabular}{p{0.8cm}p{9.4cm}p{3cm}>{\right}p{1.8cm}p{1.8cm}}
    %\begin{tabular}{llcrrr}
    \begin{tabular}{p{0.5cm}p{10cm}p{1.1cm}p{1.25cm}p{1.25cm}p{1.5cm}}
    \toprule
    \multicolumn{2}{l}{Hypothesis} & Sig$^\star$ & GT & PORL & Chen et al.\\
   \toprule
H1  & The average character length of a headline & False& 70.55 & \bf{96.21} & 73.92 \\
H2  & The average of token lengths in a headline (lower is better) & True &  4.97 & \bf{4.78} & 4.89\\
H3  & The percentage of headlines containing a question mark  & True & \bf{2.52\%} & 0.90\% & 1.19\%\\
H4  & The percentage of headlines containing a partial quote & True &11.81\%  & \bf{15.80\%} & 13.85\%\\
H5  & The percentage of headline containing full quote (lower is better) & False & 0.01\% & \bf{0.00}\% & \bf{0.00}\% \\
H6  & The percentage of headline containing signal words & True& 9.90\% & \bf{19.83\%} & 15.00\% \\
H7  & The percentage of headline containing personal or possessive pronoun & True & 28.82\% & \bf{48.67\%} & 40.35\%\\
H8  & The percentage of headline containing sentimental words & True & 68.82\% & \bf{77.40\%} & 69.37\% \\
H9  & The percentage of headline containing negative words & True& 45.09\% & \bf{52.29}\% & 44.83\% \\
H10  & The percentage of headline containing numbers & False & 20.58\% & \bf{25.22\%} & 21.06\% \\
H11  & The percentage of headline starting with personal or possessive pronoun & True & 0.64\% & \bf{1.07\%} & 0.38\% \\
    \bottomrule
    \end{tabular}
\end{table*}

\subsection{Example Comparison}
Table~\ref{tab:more_example} shows more examples for the comparison between ground-truth headline and generated headlines.

\begin{table*}[h]
\small
    \caption{Examples from the testing data showing the ground-truth headline and two generated headlines.}
    \label{tab:more_example}
    \begin{tabular}{p{5cm}p{5.5cm}p{6cm}}
        \toprule
        Ground-truth Headline & Chen et al. & \framework \\
        \toprule
        %Feral cat pictured killing and eating a wallaby 
        %& Is this the feral cat killing and eating a four-kilogram marsupial? 
        %& Is this the most threats of the country? Feral cat killing and eating a four-kilogram marsupial \\
        %\midrule
        John Bieniewicz's widow sues Bassel Saad for \$ 51m
 
        & Wife of detroit-area soccer referee filed \$ 51 million lawsuit against her husband during game last summer
        
        & Wife of soccer referee files \$ 51 million lawsuit against hot-headed player in prison for throwing a punch that killed her husband\\
        %\hline
        %\multicolumn{3}{l}{Hypothesis}\\
        \midrule 
        Chinese opera singer criticised for headdress made of feather from 80 kingfishers
        
        & Two worlds collide in china after singer posts pictures of herself wearing a headdress made from kingfisher feathers 
        
        & It's a opera! China's two singer posts pictures of herself wearing a headdress made from kingfisher feathers\\
        \midrule 
        Taylor Swift: my mom has cancer
         
        & Pop star Taylor Swift reveals her mother has cancer
         
        & `I'm a sobering': Taylor Swift reveals how her mother has terminal cancer\\
        \midrule
        Jimmy floyd hasselbaink's burton albion secure promotion to league one with victory against Morecambe

        & Burton dons 2-1 sky bet league one after Lucas Akins scores both goals
        
        & Burton Albion 2-1 Morecambe: Lucas Akins scores both goals in 2-1 win at Morecambe\\
        \midrule
        Jessica Knight suffers from pica, meaning she eats carpet underlay

        & It's the first time a child is accused of eating her parents out of house and home
        
        & Jessica Knight `can't stop eating carpet underlay and stuffing from soft furnishings'\\
    
        \bottomrule
    \end{tabular}
\end{table*}

\subsection{Implementation Details}
First, the dimension of the topic model used in our experiments is 256, and the topic model is trained by CNN/Daily Mail dataset. Second, we keep the 30000 most frequently occurring words in the training corpus. Then, we train a word2vec of 128 dimensions in the same training corpus for headline generation and use it to initialize the embedding matrix. Moreover, because the average headline length and standard deviation are 22.39 and 6.13 respectively, we truncate the input size to 30 tokens, while the article length is limited to 100 tokens, in line with previous works \cite{see2017get,chen2018fast} on CNN/Daily Mail dataset. Note that the sentence length of the input articles is not constrained when performing the testing. The length limitation of the generated headline is still maintained at 30 tokens. The LSTM model used in the abstractor and extractor are all with 256 hidden unit size, and the attention matrix size is 256. The popularity predictor is composed of a convolution neural network, an attention LSTM and a linear network to match the output dimension. The kernel sizes of the convolutional neural network for the extractor are 1, 2 and 3, and each has 300 channels. The output feature maps are concatenated, then sequentially input to the attention LSTM, which is uni-directional with 256 hidden unit size. For training, we use a mini-batch size of 32 and apply the gradient clipping by taking 2-norm for gradients and limiting gradients to 2 to avoid gradient exploding.

\bibliographystyle{aaai}
\bibliography{reference}